\documentclass[conference]{IEEEtran}
\IEEEoverridecommandlockouts
\usepackage{amsmath,amssymb,amsfonts}
\usepackage{algorithmic}
\usepackage{graphicx}
\usepackage{textcomp}
\usepackage{xcolor}

\usepackage[normalem]{ulem}
\usepackage{inconsolata}

\usepackage{tkz-graph}    
    
\usepackage{pgfplots}

\usepackage{tikz}
\usetikzlibrary{decorations.pathreplacing,calligraphy}
\usetikzlibrary{patterns}
\usetikzlibrary{arrows}
\usetikzlibrary{calc}
\usetikzlibrary{shapes}

\usepackage{dblfloatfix}
\usepackage{pgfkeys}
\usetikzlibrary{backgrounds}
\pgfplotsset{compat=newest}
\usepackage[export]{adjustbox}
    
\usepackage{comment}
    
\usepackage{xspace}

\usepackage{algorithm}
\usepackage{algorithmic,multicol}
\usepackage{multirow}
\usepackage{booktabs}

\usepackage{tabularx}

\usepackage{pdfpages}

\renewcommand{\sfdefault}{LinuxBiolinumT-LF}

\usepackage[backend=biber,
sorting=none,
doi=false,
isbn=false,
url=false,
maxnames=10,
minnames=2,
style=ieee]{biblatex}
\usepackage{caption}
\usepackage{subcaption}

\newcommand{\tool}{\textsc{T-RecX}\xspace}
\newcommand{\bresnet}{\textsc{Branchy-Resnet}\xspace}
\newcommand{\bdscnn}{\textsc{Branchy-DSCNN}\xspace}
\newcommand{\sresnet}{\textsc{SDN-Resnet}\xspace}
\newcommand{\sdscnn}{\textsc{SDN-DSCNN}\xspace}

\newcommand{\Vsubscre}[2]{$#1$\textsubscript{\textit{$#2$}\xspace}}

\definecolor{bblue}{HTML}{4F81BD}
\definecolor{rred}{HTML}{C0504D}
\definecolor{ggreen}{HTML}{9BBB59}
\definecolor{ppurple}{HTML}{9F4C7C}
\definecolor{yyellow}{HTML}{DDAE06}

\tikzstyle{bar1} = [bblue, fill=bblue, postaction={pattern=north east lines}]
\tikzstyle{bar2} = [rred, fill=rred]
\tikzstyle{bar3} = [ggreen, fill=ggreen, postaction={pattern=north west lines}]
\tikzstyle{bar4} = [ppurple, fill=ppurple, postaction={pattern=crosshatch dots}]
\tikzstyle{bar5} = [yyellow, fill=yyellow, postaction={pattern=grid}]

\newcommand{\graphHeight}{0.15\textheight}

\begin{document}

\title{\tool: Tiny-Resource Efficient Convolutional neural networks with early-eXit
}

\author{\IEEEauthorblockN{Nikhil P Ghanathe, Steve Wilton}
\textit{University of British Columbia}\\
nikhilghanathe@ece.ubc.ca
}

\maketitle


\begin{abstract}
Deploying Machine learning (ML) on milliwatt-scale edge devices (tinyML) is
gaining popularity due to recent breakthroughs in ML and Internet of Things (IoT). Most tinyML research focuses on model compression techniques that trade accuracy (and model capacity) for compact models to fit into the KB-sized tiny-edge devices. 
In this paper, we show how such models can be enhanced by the addition of an early exit intermediate classifier.   If the intermediate classifier exhibits sufficient confidence in its prediction, the network exits early thereby, resulting in considerable savings in time.  Although early exit classifiers have been proposed in previous work, these previous proposals focus on large networks, making their techniques suboptimal/impractical for tinyML applications.  
Our technique is optimized specifically for tiny-CNN sized models.
In addition, we present a method to alleviate the effect of network overthinking by leveraging the representations learned by the early exit. We evaluate \tool on three CNNs from the MLPerf tiny benchmark suite for image classification, keyword spotting and visual wake word detection tasks. Our results show that \tool 1)~improves the accuracy of baseline network, 2)~achieves 31.58\% average reduction in FLOPS in exchange for one percent accuracy across all evaluated models. Furthermore, we show that our methods consistently outperform popular prior works on the tiny-CNNs we evaluate. 

\end{abstract}


\section{Introduction}
\label{sec:intro}

Recent advancements in Machine learning (ML) and IoT have produced ML models that have compute and memory requirements low enough to be run directly
on small milliwatt-scale devices. These models enable several real-world applications such as smart farming~\cite{farmbeats}, smart cities~\cite{city1, city2}, driverless cars~\cite{car1, car2, car3}, and smart healthcare~\cite{biomedical, gesturepod}. The study of deploying such ML models on ultra-low power resource-constrained milliwatt-scale edge devices is called \textit{TinyML}~\cite{tinyml}. 

The capabilities of TinyML are limited by strict power, resource, and compute constraints of the edge devices. 
TinyML targets ultra-low-power devices (milliwatt range) with limited compute power that are often battery-operated.
TinyML models must be small enough (few kilobytes) to fit into the KB-sized tiny-edge devices and require all computations to run locally with no support from the cloud.   
Thus, one of the major objectives of TinyML is to minimize the \emph{energy per prediction}.
\begin{figure}[tb]
    \centering
    \scalebox{0.6}{\begin{tikzpicture}[font=\sffamily\small\fontfamily{LinuxBiolinumT-LF}\selectfont\large,>=stealth']

  \tikzset{ionode/.style={align=center,minimum width=1.5cm,black!75}};
  \tikzset{layernode/.style={align=center,draw,minimum height=1cm, rounded corners=3pt,inner sep=5pt,fill=black!5}};
  \tikzset{classfiernode/.style={align=center,draw,minimum height=1cm,rounded corners=4mm,inner xsep=8pt,, fill=yellow}};
  \tikzset{poolingnode/.style={align=center,draw,minimum height=1cm,rounded corners=5mm,inner xsep=8pt,fill=blue}};
  \tikzset{verticalpoolingnode/.style={align=center,draw,minimum height=2.2cm,rounded corners=5mm,inner xsep=8pt,fill=blue}};
  \tikzset{eenode/.style={align=center,draw, minimum height=2cm,rounded corners=1mm,inner xsep=8pt,fill=red}};

  \node (input) [ionode,black] {Input\\Image};
  \node (l1) at (input.south) [layernode,anchor=south,xshift=0cm, yshift=-2cm,fill=black!15] {CONV};
  \node (l2) at (l1.south) [layernode,anchor=south,xshift=0cm, yshift=-1.5cm,fill=black!15] {CONV};
  \node (l3) at (l2.south) [layernode,anchor=south,xshift=0cm, yshift=-1.5cm,fill=black!15] {CONV};
  \node (l4) at (l3.south) [layernode,anchor=south,xshift=0cm, yshift=-1.5cm,fill=black!15] {CONV};
  \node (l6) at (l4.south) [layernode,anchor=south,xshift=0cm, yshift=-1.5cm,fill=black!15] {CONV};
  \node (l7) at (l6.south) [layernode,anchor=south,xshift=0cm, yshift=-1.5cm,fill=black!15] {CONV};
  \node (ef-pool) at (l7.south) [poolingnode,anchor=south,xshift=0cm, yshift=-1.5cm, fill=blue!15] {POOL};
  \node (Cf) at (ef-pool.south) [classfiernode,anchor=south,xshift=0cm, yshift=-1.5cm] {DENSE};
  \node (out_ef) at (Cf.south) [ionode,black, xshift=0cm, yshift=-1.5cm] {Final exit};
    \node (ee-pconv) at (l2.south east) [eenode,anchor=west,xshift=1.5cm, yshift=-2cm, fill=red!15] {P\\C\\O\\N\\V};
    \node (ee-dconv) at (ee-pconv.east) [eenode,anchor=west,xshift=1cm, yshift=0cm, fill=red!15] {D\\C\\O\\N\\V};
    \node (ee-pool) at (ee-dconv.south) [poolingnode,anchor=south,xshift=0cm, yshift=-2cm, fill=blue!15] {POOL};
    \node (Ce) at (ee-pool.south) [classfiernode,anchor=south,xshift=-0cm, yshift=-1.5cm] {DENSE};
    \node (out_ee) at (Ce.south) [ionode,black, yshift=-1.5cm] {Early exit};
    
    \node (ee-layers) at (ee-pconv.north) [ionode,black, xshift=1cm, yshift=0.5cm] {Early-exit layers};
    
    \node (annotate) at (input.east) [ionode,black, align=left, xshift=4.5cm, yshift=-1.5cm] {\textbf{CONV}: \textit{Convolution}\\\textbf{POOL}: \textit{Pooling}\\\textbf{DENSE}: \textit{Fully-connected}\\\textbf{PCONV}: \textit{Pointwise CONV}\\\textbf{DCONV}: \textit{Depthwise CONV}};

     \draw [->] (input) -- (l1);
     \draw [->] (l1) -- (l2);
     \draw [->] (l2) -- (l3);
     \draw [->] (l3) -- (l4);
     \draw [->] (l4) -- (l6);
     \draw [->] (l6) -- (l7);
     \draw [->] (l7) -- (ef-pool);
     \draw [->] (ef-pool) -- (Cf);
     \draw [->] (Cf) -- (out_ef);
     
    \draw [->] (l2.east)-- ++(0, 0cm)  -- ++(0.5cm, 0)-- ++(0,-2.5cm)-- (ee-pconv.west);
    \draw [->] (ee-pconv) -- (ee-dconv);
    \draw [->] (ee-dconv) -- (ee-pool);
     \draw [->] (ee-pool) -- (Ce);
    \draw [->] (Ce) -- (out_ee);
\coordinate (a) at (ee-pconv.north west) + (-3mm, 3mm);
\coordinate (b) at (ee-dconv.north east) +(3mm, 3mm);
  \draw [dotted, thick] (ee-pconv.north west) ++ (-3mm, 3mm) --  ++(3.4cm, 0mm) -- ++(0,-3.1cm) -- ++ (-3.4cm, 0) -- ++(0, 3.1cm);
  
  
  \draw [decorate,decoration={brace,mirror, amplitude=10pt,,raise=4pt},yshift=0pt]
(l1.north west)+(-2mm,1mm) -- ++(-2mm, -2.7cm) node [black,midway,xshift=-1.3cm, yshift=3mm] {Common} node [black,midway,xshift=-1.3cm, yshift=-1mm]{block};

\draw [decorate,decoration={brace,mirror,amplitude=10pt,raise=4pt},yshift=0pt]
(l3.north west)+(-2mm,1mm) -- ++(-2mm, -10cm) node [black,midway,xshift=-1.3cm, yshift=2mm] {Final}node [black,midway,xshift=-1.3cm, yshift=-2mm]{block};

\draw [decorate,decoration={brace,  amplitude=10pt,,raise=4pt},yshift=0pt]
(ee-dconv.north east)+(4mm,1mm) -- ++(4mm, -6.6cm) node [black,midway,xshift=1.4cm, yshift=2mm] {Early-exit} node [black,midway,xshift=1.4cm, yshift=-2mm] {block};

\draw [ultra thick] (annotate.north west) +(-1mm,1mm)-- ++(5cm, 1mm) -- ++(0,-2.7cm)-- ++(-5.1cm,0) -- ++(0,2.7cm);

\end{tikzpicture}}
    \caption{CNN with early-exit}
    \label{fig:model_plus_ee}
\end{figure}
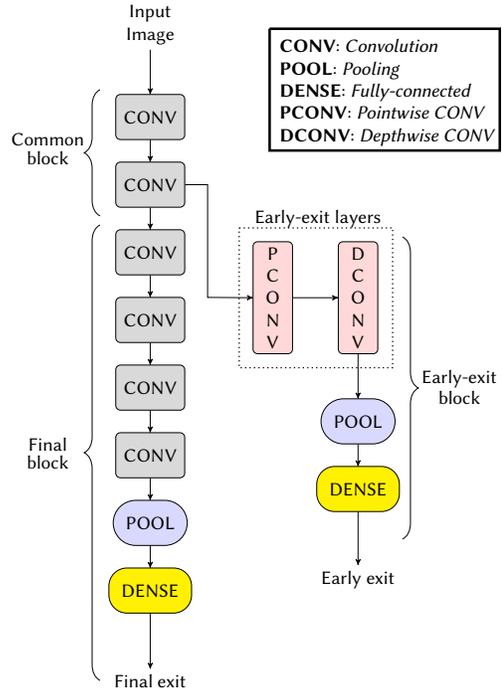
To that end, several solutions have been proposed to improve the energy-efficiency/inference time of ML models running on edge devices. The predominant body of prior work involves model compression techniques such as pruning~\cite{pruning}, quantization~\cite{quant1, quant3, quant4} and knowledge distillation~\cite{knowledge-distill}. However, compression comes at the cost of model accuracy and leads to reduced model capacity.
Other areas of work such as split-computing~\cite{split-1} and prediction cascades~\cite{idk} have been proposed for the tiny-edge setting (more details in Section~\ref{sec:related}). These methods also suffer from reduced model capacity and often require delegation to the cloud to conserve network accuracy. As a result, they are impractical for our applications.

In contrast, early-exit networks preserve the network’s model capacity while reducing the average inference time. Early-exit networks place intermediate classifiers along the depth of a baseline DNN that are capable of producing an output that approximates the output of the final classifier.
These intermediate classifiers act as additional exit paths for the DNN. If the \emph{confidence} score of the output at any intermediate classifier is above a predefined/learned threshold, the DNN exits and the output of that intermediate classifier is forwarded to the output of the DNN. Figure~\ref{fig:model_plus_ee} shows a convolution neural network (CNN) with one early-exit block. Early-exit strategies are quite general and can be applied to off-the-shelf state-of-the-art networks as a means to improve the average inference time with minimal overhead~\cite{branchynet}. In addition, early-exit can be applied in conjunction with other energy reduction techniques such as pruning and quantization. 

However, early-exits are detrimental to the network performance~\cite{msdnet}, and this negative effect worsens as model size decreases (Section~\ref{sec:motivation}). Our key contributions are summarized as follows. 
\begin{itemize}
    \item We identify key challenges in applying early-exits for tiny-CNNs (Section~\ref{sec:motivation}), and demonstrate that prior early-exit works are inefficient/ineffective (Section~\ref{subsec:comp_prior}).
    \item We propose \tool, an early-exit architecture specialized for tiny-CNNs that addresses these challenges (Section~\ref{sec:design}).
    \item We develop a method to mitigate the effect of network \emph{overthinking}~\cite{sdn} for tiny-CNNs that reclaims some of the lost accuracy due to early-exit (Section~\ref{subsec:ee-view}).
    \item \tool achieves 31.58\% average reduction in FLOPS for one percent accuracy trade-off across all the CNNs we evaluate from the MLPerf tiny benchmark suite~\cite{tiny-benchmark}.
    \item Our methods increases the accuracy of the baseline network in all tiny-CNNs we evaluate.
\end{itemize}
To the best of our knowledge, we are the first to propose the use of an early-exit strategy for tiny-CNNs that optimizes for both accuracy and floating-point operations per second (FLOPS) reduction.

\section{Related Work}
\label{sec:related}

\noindent\textbf{Resource efficient machine learning}~Several works have been proposed for improving the average inference time of CNNs at the edge. The primary methods relevant for the tinyML space involve model compression techniques~\cite{compression-1, quant1, quant3, knowledge-distill, pruning} to fit models into tiny-edge devices. Other approaches propose changes to network topology and explore resource and computationally efficient versions of traditional neural network models~\cite{fastgrnn, scnn, emirnn, Seedot, mafia}. These efforts are complementary to our work, and can be used in combination with our proposed early-exit technique.
Two other solutions, which are closer to our approach,
are split computing~\cite{split-1, split-2, split-3} and prediction cascades~\cite{idk, cascade-1, cascade-2, cascade-3}. Split computing seeks to partition the ML model between the edge device and cloud. 
Cascade networks consist of a cascade of DNN models that get progressively larger to obtain a prediction cascade. Unlike split computing, the computations are not reused in cascading networks. Since both approaches rely on the cloud, they are impractical for tinyML.

\vspace{1mm}\noindent\textbf{Early-exit networks}~Teerapittayanon et al.~\cite{branchynet} first introduced early-exit strategies for CNNs. That work proposes adding two early-exits and uses multiple convolution blocks as part of its early-exit architecture. Similarly, Szegedy et al.~\cite{conv-at-ee} use convolution layers in its early-exit blocks. 
Another relevant work, Multiscale Densenet (MSDNet)~\cite{msdnet} generates multiple feature maps after each layer with different scales. The authors show that adding early-exits interferes with the features learnt at the later layers. Hence, the multi-scale feature maps help in maintaining coarse-level features throughout the network, which helps the accuracy of early-exit classifiers. The early-exit blocks consists of one average pooling layer followed by a dense layer. 
Bonato et al.~\cite{class-spec} focuses on boosting the classification rate of a specific class at early-exit. Its early-exit architecture is identical to that of MSDNet. The overall accuracy of the network is improved since each early-exit block has access to all feature maps from preceding layers. Skipnet~\cite{skipnet} selectively skips convolution layers during inference by using a small gating network without sacrificing accuracy. Shallow deep network~\cite{sdn} introduce early-exits to CNNs to study the problem of network overthinking and mitigate its destructive effect. It uses multiple early-exits with pooling and dense layers. 
Leontiadis et al.~\cite{always-personal} describes hierarchical models from a global base model for the mobile edge. The appropriate model is selected at runtime depending on the computation budget.

Several other works~\cite{ee-1, hydranet, gaternet, ee-2, ee-3, ee-4} present early-exit networks. Some tinyNN-specific early-exit works~\cite{energy-aware-EE-1, energy-aware-EE-2, energy-aware-EE-3} exit purely based on energy (battery) constraints and sacrifice heavily on accuracy. In our work, we optimize for both accuracy and FLOPS and provide a tunable parameter that can trade-off between both. Almost all prior works include multiple early-exit classifiers for balancing the network accuracy, which is unsustainable for tiny-CNNs. Furthermore, we find that apart from the works in~\cite{class-spec} and~\cite{msdnet}, none explore utilizing high-level representations learned by the earlier layers at the final layers. Unlike \tool, both ~\cite{class-spec} and~\cite{msdnet} leverage feature maps from multiple exit points, which results in a prohibitive cost for the tiny-edge environment (Section~\ref{subsec:ee-view}).

\section{Motivation and Challenges}
\label{sec:motivation}
Deep neural networks are growing increasingly deeper in the pursuit of higher accuracy. However, the accuracy gains by using deeper networks are paltry after a certain point~\cite{idk}. For example, a Resnet50~\cite{resnetv1} model for an image classification task on the ImageNet dataset~\cite{imagenet} achieves an accuracy of 76.2\%, but, a Resnet101 model (double the layers) for the same task achieves an accuracy of 77.4\%. It is evident that a smaller model will suffice for many applications.
Previous works~\cite{idk, branchynet,msdnet,class-spec} have demonstrated that there are a large number of \emph{easy-to-classify} samples that do not require the full depth of the DNNs. Early-exit networks exploit this property to improve the average inference time of the network. However, as described in Section~\ref{sec:related}, most of the contemporary works target larger neural networks that are not suitable for deployment on tiny-edge devices. In this section, we enumerate the major challenges in applying the
early-exit technique on tiny-CNNs.

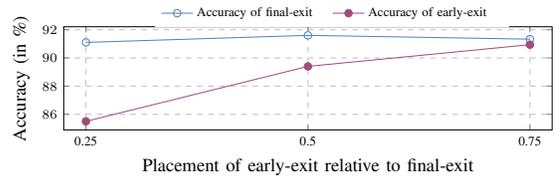
\begin{figure}[tb]
    \centering
    \scalebox{0.7}{\begin{tikzpicture}
	\pgfplotstableread{
	ee-placement 	ee-accuracy ef-accuracy
	0.25               85.50         91.1
	0.5                89.40         91.60
	0.75               90.94         91.33

	}\data
		\begin{axis}[
			width=\columnwidth+2cm,
			height=\graphHeight,
			%
			major tick length=3pt,
			%
			major grid style={dashed,color=gray!50},
			minor grid style={color=gray!50},
			ymajorgrids=true,
			xmajorgrids=true,
			xminorgrids=true,
			%
			xtick=data,
			xticklabels from table={\data}{ee-placement},
			xticklabel style={text height=5pt,font=\scriptsize},
			%
			xlabel = Placement of early-exit relative to final-exit,
			xlabel style={font=\scriptsize},
		    xlabel near ticks,
			%
			yticklabel style={font=\scriptsize},
			%
			ylabel style={align=center,font=\scriptsize},
			ylabel=Accuracy (in \%),
			ylabel near ticks,
			legend style={at={(0.55,1.0)},anchor=south,cells={align=left},draw=none,font=\scriptsize,text width=2.5cm},
			legend columns=-1,
			%
			enlarge x limits=0.05,
			%
		]

		\addplot [bblue, mark=o] table[x=ee-placement,y=ef-accuracy] {\data};
		\addplot [ppurple, mark=*] table[x=ee-placement,y=ee-accuracy] {\data};
		
		
		\addlegendentry{Accuracy of final-exit}
		\addlegendentry{Accuracy of early-exit}

		\end{axis}
	\end{tikzpicture}}
    \caption{The accuracy of final classifier with respect to the relative location of early-exit.\vspace{-0.4cm}}
    \label{fig:final_accuracy_drop}
\end{figure}

\vspace{1mm}
\noindent
\textbf{Additional exit-path affects network accuracy}
TinyML models are primarily optimized to
reduce their memory and compute requirements through aggressive model compression~\cite{pruning, knowledge-distill, quant4,cnn-quant-1} to fit on low-end devices while compromising on accuracy to some degree. Consequently,
the compressed models are stripped to their bare-bones and have a high degree of \emph{neuron co-adaptations} between successive layers compared to their larger counterparts. Any change to the model architecture, like adding an early-exit is adversarial and detriments the performance of the network. Additionally, given the size of tiny-CNNs, the early-exit block has to be attached as close to the input as possible to see any real performance gains, which irreversibly disrupts the crucial initial layers of the CNN~\cite{transfer-learning}. 
Figure~\ref{fig:final_accuracy_drop} illustrates this effect for a 4-layer depthwise separable CNN model (DS-CNN). Figure~\ref{fig:final_accuracy_drop} shows the standalone accuracy of the final exit of the DS-CNN~\cite{DS-CNN} model when a single early-exit block is attached at $\frac{1}{4}$\textsuperscript{th}, half and $\frac{3}{4}$\textsuperscript{th} of the network. The standalone accuracy of final exit refers to the accuracy of the final classifier when all inputs take the final exit (no early-exit). The standalone accuracy of the early-exit refers to the accuracy when all samples take the early-exit.
From the figure, we observe that the configuration where the early-exit is attached at $\frac{1}{4}$\textsuperscript{th} of the network i.e. closest to the input results in the highest harm to the standalone accuracy of the final classifier for the reasons discussed above. 

\vspace{1mm}
\noindent
\textbf{Existing early-exit strategies are ineffective for tiny-CNNs}
The predominant body of existing work on early-exits target large neural networks, which also suffer from accuracy degradation with even a single early-exit classifier~\cite{msdnet}.
To combat this challenge, a majority of the existing works attach multiple early-exit blocks along the length of the CNN~\cite{sdn, branchynet, class-spec, why-ee} to avoid the final-exit as much as possible. For example, early-exit networks from Kaya et. al.~\cite{sdn} place their early-exit classifiers at 15\%, 30\%, 45\%, 60\%, 75\% and 90\% of the Resnet-56 network for image classification on the CIFAR-10 dataset~\cite{cifar-10}. For the same task, Teerapittayanon et. al.~\cite{branchynet} adds two early-exit branches after the 2\textsuperscript{nd} convolution layer and 37\textsuperscript{th} convolution layer of the Resnet-110 network. In contrast, the tiny-CNN model we evaluate for the same task and dataset is a tiny Resnet-8 model~\cite{tiny-benchmark}. Furthermore, the large CNN networks targeted in prior works are less vulnerable to an extra output compared to tiny-CNNs because the larger networks have enough layers after the early-exit block to relearn the complex neuron relations disturbed by the early-exit. Moreover, the relative placement of the early-exit in larger networks usually bypasses the initial crucial layers. 
Another approach adopted by prior works~\cite{msdnet, class-spec} is to combine feature maps from all previous early-exits. However, the cost incurred by both placing multiple early-exits and combining feature maps from previous layers is untenable for tiny-CNNs.
For example, the Resnet-8 model with two early-exits results in 17.2\% increase in model parameters, which is huge from a tinyML perspective. As a result, tiny-CNNs are limited to a single early exit placed as close to the input as possible because the inference cost (FLOPS) quickly grows beyond that of the unmodified baseline network as we move away from the input. 
Unfortunately, the proximity of the early-exit to the input in CNNs results in notable accuracy degradation.
Thus, early-exit in tiny-CNNs induce a trade-off between network accuracy and FLOPS. The primary contribution of this work is to study the feasibility and effects of early-exit on tiny-CNNs and formulate techniques that balances the overall network accuracy while delivering notable reduction in FLOPS.

\section{\tool}
\label{sec:design}
\tool modifies the baseline tiny-CNN model by adding a single early-exit block as shown in Figure~\ref{fig:model_plus_ee}. 
However, the reduced model capacity of the early-exit classifier induces a high scope for misclassification. 
\tool addresses this challenge using a novel early-exit block for tiny-CNNs and employing a joint-training method.
We formulate the problem formally in Section~\ref{subsec:prob}. Section~\ref{subsec:arch-ee} describes the architecture of the early-exit block and a training method with a joint loss function. Section~\ref{subsec:ee-view} describes a method to mitigate the effect of network \emph{overthinking} at the final classifier by utilizing information from the early-exit block, thereby enhancing the overall network accuracy.

\subsection{Problem Setup}
\label{subsec:prob}
Given a baseline trained tiny-CNN model \Vsubscre{M}{B}($x$, $\theta$), \tool adds an early-exit block \Vsubscre{E}{e} as shown in Figure~\ref{fig:model_plus_ee}. $\theta$ represents the model parameters learned after training. The model after adding the early-exit block is denoted by \Vsubscre{M}{B\textsubscript{$e$}}($x$, $\theta$\textsubscript{$e$}) and is defined as, 
\begin{equation}
    M\textsubscript{B\textsubscript{e}}(x, \theta_e)   = 
\begin{cases}
    argmax(z_i\textsubscript{:E{\text-}e}),& \text{if } max(z_i\textsubscript{:E{\text-}e})\geq \rho\\
    argmax(z_i\textsubscript{:E{\text-}f}),              & \text{otherwise}
\end{cases}\label{eq:model_ee}
\end{equation}
\noindent 
where $z_i\textsubscript{:E{\text-}e}$ and $z_i\textsubscript{:E{\text-}f}$ are the softmax scores of the early-exit classifier and the final classifier respectively computed from the $i\textsuperscript{th}$ input sample $x_i$. $\rho$ is the \emph{confidence threshold} such that $0\leq\rho\leq1$ , which can be used to tune the early-exit rate (\Vsubscre{EE}{rate}) to tradeoff between classification accuracy and network inference cost. The softmax score at early-exit is used as the confidence metric. If the highest score of the softmax output i.e. the score of the predicted label at the early-exit is greater than or equal to a threshold ($\rho$), the CNN exits at the early-exit and the prediction of the early-exit classifier is forwarded to the output of the CNN. Otherwise, the prediction of the final classifier is forwarded to the output. 

The total accuracy of the model is given by, 
\begin{small}
\begin{eqnarray}
\textrm{acc}\textsubscript{total} & = & \frac{\sum_{\forall i\in \mathcal{N}_e}^{}(y_i==\hat{y_i}) + \sum_{\forall i\in \mathcal{N}_f}^{}(y_i==\hat{y_i})}{\eta_e +\eta_f}\label{eq:acc_total}
\end{eqnarray}
\end{small}
\noindent
where $\mathcal{N}_e$ and $\mathcal{N}_f$ are the set of input samples that exit at the early-exit (\Vsubscre{E}{e}) and the final exit (\Vsubscre{E}{f}) and, $\eta_e$ and $\eta_f$ are the number of test samples present in $\mathcal{N}_e$ and $\mathcal{N}_f$ respectively. The total number of test samples ($\eta$) is given by $\eta$ = $\eta_e + \eta_f$. $y_i$ and $\hat{y_i}$ denote the predicted label and the true label respectively.

At a high level, the objective of \tool is to maximize both the early-exit rate given by $\textrm{\Vsubscre{EE}{rate}}(\rho) =  \frac{\eta_e}{\eta_e+\eta_f}$, and overall accuracy of the model given by Eq~\ref{eq:acc_total}. 
The \Vsubscre{EE}{rate}($\rho$) is increased by lowering the threshold $\rho$. However, the accuracy at early-exit degrades quickly as the \Vsubscre{EE}{rate} increases beyond a certain point because of the limited model capacity of the early-exit classifier. On the other hand, as the \Vsubscre{EE}{rate} is reduced, the network inference cost ($\mathcal{C}$) increases. We describe the network inference cost in terms of FLOPS.  
As shown in Figure~\ref{fig:model_plus_ee}, the early-exit model \Vsubscre{M}{B\textsubscript{$e$}}($\theta$\textsubscript{$e$}) can be divided into three parts i.e. common block, early-exit block and the final block. The network inference cost is given by,
\begin{equation}
    \mathcal{C}  = 
\begin{cases}
    \mathcal{C}\textsubscript{E{\text-}e},&     \text{if early-exit}\\
    \mathcal{C}\textsubscript{E{\text-}f},&     \text{otherwise}
\end{cases}\label{eq:cost}
\end{equation}
where, $\mathcal{C}\textsubscript{E{\text-}e} = \mathcal{C}\textsubscript{common} + \mathcal{C}\textsubscript{EE}$ and $\mathcal{C}\textsubscript{E{\text-}f} = \mathcal{C}\textsubscript{E{\text-}e} + \mathcal{C}\textsubscript{final}$

$\mathcal{C}$\textsubscript{common}, $\mathcal{C}$\textsubscript{EE}
and $\mathcal{C}$\textsubscript{final} are the inference cost of the common block, early-exit block and the final block of the network respectively. $\mathcal{C}\textsubscript{E{\text-}e}$ and $\mathcal{C}\textsubscript{E{\text-}f}$ give the inference cost of the early-exit-classifier and the final classifier. The objective of \tool distills to minimizing the total network inference cost ($\mathcal{C}$) with an acceptable trade-off on classification accuracy.



\subsection{Architecture of Early-exit block and Training Methodology}
\label{subsec:arch-ee}
Since the feature map input to the early-exit block is a high-dimensional variable, it is not highly informative of its true label~\cite{eefmpas_high_dim}. As a result, 
most prior works only have a feature reduction stage (pooling) before the classification (dense) layer. Our preliminary experiments showed that for tiny-CNNs, na\"{\i}ve feature reduction leads to massive information loss and accuracy degradation. 
Further investigations revealed that high-dimensional feature maps do not posses sparse class-specific information that help in lowering network confusion (ambiguity in prediction).
Our analysis showed that the early-exit classifier requires additional information to resolve \emph{confusion}. To accomplish this, we increase the number of channels by introducing a combination of \emph{pointwise} and \emph{depthwise} convolutions layers.
Figure~\ref{fig:model_plus_ee} shows the architecture of \tool's early-exit block. The early-exit block consists of a pointwise convolution that increases the number of channels followed by a depthwise convolution that extracts feature information from each channel. By increasing the number of channels, we increase the \emph{perceptions} of the input image thereby lowering the \emph{confusion} of the early-exit classifier. Simultaneously, we increase the stride of the depthwise convolution layer to achieve partial feature reduction before the pooling and dense (classification) layers. Our analyses showed that the accuracy of the early-exit steadily increases as its channel width expands, as long as the channel width of the early-exit is below or equal to that of the final exit. The accuracy gain saturates quickly beyond this point. We find that the optimal configuration in terms of both accuracy and resource usage is when the PCONV layer of the early-exit doubles the number of channels at its input. A higher channel width incurs significant memory and compute overhead whereas, a lower channel width results in meager accuracy gains. In our evaluations, we configure the PCONV layer for the same. 

\vspace{1mm}
\noindent
\textbf{Training Methodology}
Given an untrained baseline model \Vsubscre{M}{B\textsubscript{e}} attached with an early-exit block as shown in Figure~\ref{fig:model_plus_ee}, \tool trains the network parameters $\theta_e$, such that the network inference cost $\mathcal{C}$ is minimized with an acceptable trade-off on accuracy. \tool uses the cross-entropy loss with the Adam optimizer~\cite{adam} as the objective function, which is the same as that used by the reference models~\cite{tiny-benchmark}.
The early-exit block is added to the baseline network and all the weights are learned simultaneously. 
Since the early-exit CNN is a dual-output neural network, the combined loss value is obtained by adding the loss terms from each output. The proximity of the early-exit to the input layer results in the weights of the initial layers (common block) being heavily optimized for the early-exit leading to accuracy drop at the final classifier (Section~\ref{sec:motivation}). Therefore, to attenuate the effect of the early-exit on the common block layers, \tool weights the loss term at the early-exit by a factor $w\textsubscript{E{\text-}e}$ such that $0<w\textsubscript{E{\text-}e}<1$. 
Thus, the total loss is given by, $
\mathcal{L}= w\textsubscript{E{\text-}e}\cdot\mathcal{L}\textsubscript{E{\text-}e} + \mathcal{L}\textsubscript{E{\text-}f}$. The loss terms computed at the early and final exit are $\mathcal{L}\textsubscript{E{\text-}e}$ and $\mathcal{L}\textsubscript{E{\text-}f}$ respectively. 
A smaller weighting of the loss function at early-exit reduces the impact on the co-adaptations of neurons in the common block of the network. 
We experiment with various values of $w\textsubscript{E{\text-}e}$ for each network we evaluate. The $w\textsubscript{E{\text-}e}$ value that gives the maximum standalone accuracy at the final classifier i.e. the value of $w\textsubscript{E{\text-}e}$ that causes minimum disturbance in the complex relations learned between the layers of the common block and the final block is selected. The values that satisfy this constraint are reported in Section~\ref{sec:methodology}. 

\subsection{Early-view assisted classification}
\label{subsec:ee-view}
Despite the effectiveness of the proposed early-exit block and the training method, there are still some samples that are misclassified at the early-exit but would have been correctly classified at the final exit. Similarly, there exists a set of samples that would have been correctly classified at the early-exit had they exited there. The final classifier misclassifies these samples due to \emph{overthinking}. From Kaya et al.~\cite{sdn}, a network is said to \emph{overthink} when 1)~the higher-level representations of an input sample computed by an early layer suffices for a correct classification and 2)~further computation by the subsequent layers lead to misclassification by the final classifier. Hence, the biggest challenge with adding a single early-exit is to determine which input samples should exit early, and which should not. At best, this can be predicted by a heuristic, inevitably leading to accuracy loss. This is an open problem and almost all solutions to tackle it involve multiple early-exits. Although, the problem of overthinking in tiny-CNNs is not as dominant as in large CNNs, we observe that the problem still persists. Early-exit networks provide an opportunity to mitigate the effect of overthinking and reclaim some of the lost accuracy.
Since the early-exit block is computed before the final classifier, it presents us with an opportunity to leverage information from the early-exit block to improve the performance of the final classifier. However, na\"{\i}ve concatenation of the output of the early-exit block with the final layer yields poor results. \tool recognizes that the filter weights of the convolution layers in the early-exit block capture the essence of the features that are learnt at the early-exit.   

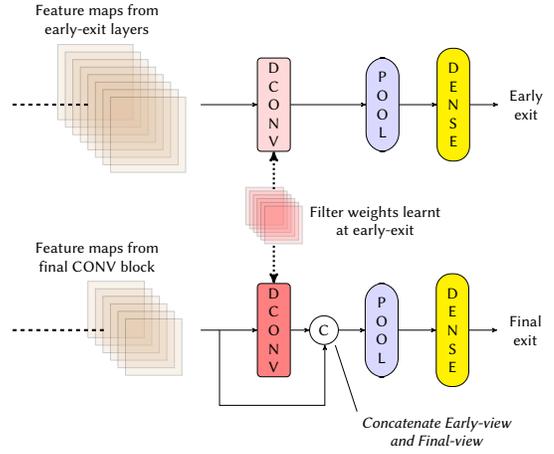
\begin{figure}[t]
    \centering
    \scalebox{0.5}{\begin{tikzpicture}[font=\sffamily\small\fontfamily{LinuxBiolinumT-LF}\selectfont\large,>=stealth']

  \tikzset{ionode/.style={align=center,minimum width=1.5cm,black!75}};
  \tikzset{layernode/.style={align=center,draw,minimum height=2cm, rounded corners=3pt,inner sep=5pt,fill=black!5}};
  \tikzset{classfiernode/.style={align=center,draw,minimum height=3cm,rounded corners=4mm,inner xsep=8pt,, fill=yellow}};
  \tikzset{poolingnode/.style={align=center,draw,minimum height=2.5cm,rounded corners=5mm,inner xsep=8pt,fill=blue}};
  \tikzset{circlenode/.style={align=center,draw,minimum height=0.5cm,rounded corners=5mm,inner xsep=4pt,fill=blue}};
   \tikzset{eenode/.style={align=center,draw, minimum height=2cm,rounded corners=1mm,inner xsep=8pt,fill=red}};

  \node (anchor) [ionode,black] {};
  \node (input_fmaps) at (anchor.north) [ionode, black, xshift=2.5cm, yshift=1.1cm] {Feature maps from\\early-exit layers};
  \node (input_weights) at (anchor.south) [ionode, black, xshift=9.9cm, yshift=-4cm] {Filter weights learnt\\at early-exit};
  \node (ee-dconv) at (anchor.east) [eenode,anchor=west,xshift=6cm, yshift=-1cm, fill=red!15] {D\\C\\O\\N\\V};
  \node (ee-pool) at (ee-dconv.east) [poolingnode,anchor=west,xshift=2cm, yshift=0cm, fill=blue!15] {P\\O\\O\\L};
  \node (Ce) at (ee-pool.east) [classfiernode,anchor=west,xshift=1cm] {D\\E\\N\\S\\E};  
  \node (out_ee) at (Ce.east) [ionode,black, xshift=1.5cm] {Early\\exit};


  \node (ef-dconv) at (ee-dconv.south) [eenode,anchor=south,xshift=0cm, yshift=-6cm, fill=red!50] {D\\C\\O\\N\\V};
  \node (concat) at (ef-dconv.east) [circlenode, circle, anchor=west,xshift=0.5cm, yshift=0cm, fill=white!15] {C};
  \node (ef-pool) at (concat.east) [poolingnode,anchor=west,xshift=0.7cm, yshift=0cm, fill=blue!15] {P\\O\\O\\L};
  \node (Cf) at (ef-pool.east) [classfiernode,anchor=west,xshift=1cm] {D\\E\\N\\S\\E};
  \node (out_ef) at (Cf.east) [ionode,black, xshift=1.5cm] {Final\\exit};
  
  \node (input_fmaps_ef) at (input_fmaps.south) [ionode, black, xshift=0cm, yshift=-5.7cm] {Feature maps from\\final CONV block};
  
  \node (concat_annotation) at (concat.south) [ionode,black, xshift=3cm, yshift=-2.3cm] {\textit{Concatenate Early-view}\\\textit{and Final-view}};
  
  \draw[fill=brown!50,opacity=0.2](anchor.east) ++ (0.7cm,-1.4cm) -- ++(2cm,0)-- ++(0,2cm) -- ++(-2cm,0cm)-- ++(0, -2cm);
  \draw[fill=brown!50,opacity=0.2](anchor.east) ++ (0.9cm,-1.6cm) -- ++(2cm,0)-- ++(0,2cm) -- ++(-2cm,0cm)-- ++(0, -2cm);
  \draw[fill=brown!50,opacity=0.2](anchor.east) ++ (1.1cm,-1.8cm) -- ++(2cm,0)-- ++(0,2cm) -- ++(-2cm,0cm)-- ++(0, -2cm);
  \draw[fill=brown!50,opacity=0.2](anchor.east) ++ (1.3cm,-2cm) -- ++(2cm,0)-- ++(0,2cm) -- ++(-2cm,0cm)-- ++(0, -2cm);
  \draw[fill=brown!50,opacity=0.2](anchor.east) ++ (1.5cm,-2.2cm) -- ++(2cm,0)-- ++(0,2cm) -- ++(-2cm,0cm)-- ++(0, -2cm);
  \draw[fill=brown!50,opacity=0.2](anchor.east) ++ (1.7cm,-2.4cm) -- ++(2cm,0)-- ++(0,2cm) -- ++(-2cm,0cm)-- ++(0, -2cm);
  \draw[fill=brown!50,opacity=0.2](anchor.east) ++ (1.9cm,-2.6cm) -- ++(2cm,0)-- ++(0,2cm) -- ++(-2cm,0cm)-- ++(0, -2cm);
  \draw[fill=brown!50,opacity=0.2](anchor.east) ++ (2.1cm,-2.8cm) -- ++(2cm,0)-- ++(0,2cm) -- ++(-2cm,0cm)-- ++(0, -2cm);
   \draw[fill=red!50,opacity=0.2](anchor.east) ++ (5.7cm,-4.2cm) -- ++(1cm,0)-- ++(0,1cm) -- ++(-1cm,0cm)-- ++(0, -1cm);
   \draw[fill=red!50,opacity=0.2](anchor.east) ++ (5.8cm,-4.3cm) -- ++(1cm,0)-- ++(0,1cm) -- ++(-1cm,0cm)-- ++(0, -1cm);
   \draw[fill=red!50,opacity=0.2](anchor.east) ++ (5.9cm,-4.4cm) -- ++(1cm,0)-- ++(0,1cm) -- ++(-1cm,0cm)-- ++(0, -1cm);
   \draw[fill=red!50,opacity=0.2](anchor.east) ++ (6cm,-4.5cm) -- ++(1cm,0)-- ++(0,1cm) -- ++(-1cm,0cm)-- ++(0, -1cm);
   \draw[fill=red!50,opacity=0.2](anchor.east) ++ (6.1cm,-4.6cm) -- ++(1cm,0)-- ++(0,1cm) -- ++(-1cm,0cm)-- ++(0, -1cm);
   \draw[fill=red!50,opacity=0.2](anchor.east) ++ (6.2cm,-4.7cm) -- ++(1cm,0)-- ++(0,1cm) -- ++(-1cm,0cm)-- ++(0, -1cm);

  \draw[fill=brown!50,opacity=0.2](anchor.east) ++ (1.5cm,-7.2cm) -- ++(1.5cm,0)-- ++(0,1.5cm) -- ++(-1.5cm,0cm)-- ++(0, -1.5cm);
  \draw[fill=brown!50,opacity=0.2](anchor.east) ++ (1.7cm,-7.4cm) -- ++(1.5cm,0)-- ++(0,1.5cm) -- ++(-1.5cm,0cm)-- ++(0, -1.5cm);
  \draw[fill=brown!50,opacity=0.2](anchor.east) ++ (1.9cm,-7.6cm) -- ++(1.5cm,0)-- ++(0,1.5cm) -- ++(-1.5cm,0cm)-- ++(0, -1.5cm);
  \draw[fill=brown!50,opacity=0.2](anchor.east) ++ (2.1cm,-7.8cm) -- ++(1.5cm,0)-- ++(0,1.5cm) -- ++(-1.5cm,0cm)-- ++(0, -1.5cm);
  \draw[fill=brown!50,opacity=0.2](anchor.east) ++ (2.3cm,-8cm) -- ++(1.5cm,0)-- ++(0,1.5cm) -- ++(-1.5cm,0cm)-- ++(0, -1.5cm);
  \draw[fill=brown!50,opacity=0.2](anchor.east) ++ (2.5cm,-8.2cm) -- ++(1.5cm,0)-- ++(0,1.5cm) -- ++(-1.5cm,0cm)-- ++(0, -1.5cm);
  
  \draw[->] (ee-dconv) -- (ee-pool);
  \draw[->] (ee-pool) -- (Ce);
  \draw[->] (Ce) -- (out_ee);
  
  \draw[->] (ef-dconv) -- (concat);
  \draw[->] (concat)-- (ef-pool);
  \draw[->] (ef-pool) -- (Cf);
  \draw[->] (Cf) -- (out_ef);
  
  \draw[->, dotted, ultra thick]  (ee-dconv.south) ++ (0cm,-1cm) -- (ee-dconv.south); 
  \draw[->, dotted, ultra thick]  (ef-dconv.north) ++ (0cm,1cm) -- (ef-dconv.north); 

  \draw[->]  (ee-dconv.west) ++ (-1.5cm,0cm) -- (ee-dconv.west); 
  \draw[->]  (ef-dconv.west) ++ (-1.5cm,0cm) -- (ef-dconv.west); 
  \draw[->]  (ef-dconv.west) ++ (-1cm,0cm) -- ++(0, -2cm) -- ++(2.8cm, 0) -- ++(0,1.6cm);
  \draw[dashed, ultra thick]  (ee-dconv.west) ++ (-6.5cm,0cm) -- ++(2cm, 0); 
  \draw[dashed, ultra thick]  (ef-dconv.west) ++ (-6.5cm,0cm) -- ++(2.5cm, 0); 
  \draw (concat) ++(0.3cm,-0.4cm)  -- (concat_annotation.north west);

\end{tikzpicture}}
    \caption{Early-view assisted classification}
    \label{fig:early-view}
\end{figure}

A convolution filter extracts certain features from the input depending on the values of the filter weights. For example, a Sobel filter~\cite{sobel} detects edges whereas a Laplacian filter~\cite{laplacian} detects blobs. Similarly, the filter weights learned by each convolution layer in a CNN extracts relevant features in a high-dimensional space. The filters weights learned by the convolution layers in the earlier layers are adept at extracting higher-level features like shapes, appearance and outlines. On the other hand, the convolution layers closer to the final classifier focus more on complex features. Occasionally, the eminence of higher-level features learnt by the earlier layers diminishes at the final classifier leading to overthinking. \tool addresses this problem by enhancing the higher-level features at the final exit. Figure~\ref{fig:early-view} illustrates the setup for early-view assisted classification. \tool assists the final classifier in making its prediction by providing an \emph{early-view} of the input image as follows.

\vspace{1mm}\noindent \textit{1)}~\tool makes an identical copy of the \emph{depthwise convolution} layer ($DCONV$\textsubscript{E-e}) from the early-exit block. The newly created depthwise convolution layer is denoted as $DCONV$\textsubscript{E-f}.

\vspace{1mm}\noindent \textit{2)}~The input to $DCONV$\textsubscript{E-f} are the feature maps generated by the final convolution block of the CNN.

\vspace{1mm}\noindent \textit{3)}~The output of $DCONV$\textsubscript{E-f} is concatenated with the output of the final convolution block, which is subsequently fed into the final classifier (dense layer) for a prediction.

During training, the filter weights learned by $DCONV$\textsubscript{E-e} are copied over to $DCONV$\textsubscript{E-f} at the end of every training batch for approximately the first 80\% of the training epochs. For the final 20\% of the epochs, the filter weights of $DCONV$\textsubscript{E-f} are free to tune itself to learn the co-adaptations required for assisting the final classifier.

Thus, the $DCONV$\textsubscript{E-f} layer convolutes the output of the final convolution block using the filter weights learned at the early-exit. Since the filter weights learned by the earlier convolution layers provide a different (high-level) \emph{view} of the input image, concatenating this view with that of the final convolution layer allays some confusion in the final classifier caused due to overthinking. The early-view (EV) feature maps thus obtained act as a \emph{regularizer} for the final dense layer and mitigates the negative effect of overthinking. Therefore, the accuracy of the final classifier improves thereby, increasing the overall accuracy of the early-exit network. However, our preliminary evaluations revealed that excessive channel width of $DCONV$\textsubscript{E-f} results in accuracy degradation because the early-view overpowers the final-view. Thus, the channel width must be chosen such that the early-view aids the final-view instead of suppressing it. Our empirical evaluations indicate that the maximum performance gains are obtained when 1)~the channel width of $DCONV$\textsubscript{E-f} is less than half the channel width of the final CONV block, and 2)~the channel width of $DCONV$\textsubscript{E-f} is at least half the channel width of the early-exit block. To optimize for memory, in our evaluations, we set the channel width of $DCONV$\textsubscript{E-f} to be half the channel width of the early-exit. Section~\ref{subsec:results-ee-view} discusses the effectiveness of this approach.

\begin{figure*}[hbt!]
    \centering
    \scalebox{0.9}{
    \hspace{-0.4cm}
    \begin{subfigure}[]{0.3\paperwidth}
        \centering
        \begin{tikzpicture}
	\pgfplotstableread{
	ref-accuracy 	ref-flops
	87.2            25.28

	}\data
		\begin{axis}[
			width=\paperwidth/3,
			height=\graphHeight*2 - 1.8cm,
			%
			major tick length=3pt,
			%
			major grid style={dashed,color=gray!50},
			minor grid style={color=gray!50},
			ymajorgrids=true,
			xmajorgrids=true,
			%
			xticklabel style={text height=5pt,font=\scriptsize},
			%
			xlabel = Accuracy(\%),
			xlabel style={font=\scriptsize},
		    xlabel near ticks,
                xmin = 78,
			%
			yticklabel style={font=\scriptsize},
			%
			ylabel style={align=center,font=\scriptsize},
			ylabel=FLOPS (\textit{in millions}),
			ylabel near ticks,
			legend style={at={(0.55,1.0)},anchor=south,cells={align=left},draw=none,font=\scriptsize,text width=2.5cm},
			legend columns=-1,
			%
			enlarge x limits=0.1,
			%
		]

		\addplot [only marks, bblue, mark=*] table[x=accuracy, y=flops] {./data/benefit_curve_data_IC_mv.dat};
		\addplot [only marks, ppurple, mark=o] table[x=accuracy, y=flops] {./data/benefit_curve_data_IC_no_mv.dat};
		\addplot [only marks, black, solid,thick, mark=x] table[x=ref-accuracy, y=ref-flops] {\data};
		 \draw [loosely dashed, thick, red] (87.2,0) -- (87.2,25.28) -- (0,25.28);
		
		
		\addlegendentry{with early-view}
		\addlegendentry{without early-view}

        
		\end{axis}
	\end{tikzpicture}
        \label{fig:benefit_IC}
        \caption{Resnet}
    \end{subfigure}
    \hspace{0.6cm}
    \begin{subfigure}[]{0.3\paperwidth}
        \centering
        \begin{tikzpicture}
	\pgfplotstableread{
	ref-accuracy 	ref-flops
	92.2            5.54

	}\data
		\begin{axis}[
			width=\paperwidth/3,
			height=\graphHeight*2- 1.8cm,
			%
			major tick length=3pt,
			%
			major grid style={dashed,color=gray!50},
			minor grid style={color=gray!50},
			ymajorgrids=true,
			xmajorgrids=true,
			%
			xticklabel style={text height=5pt,font=\scriptsize},
			%
			xlabel = Accuracy(\%),
			xlabel style={font=\scriptsize},
		    xlabel near ticks,
			%
			yticklabel style={font=\scriptsize},
			%
			ylabel style={align=center,font=\scriptsize},
			ylabel near ticks,
			legend style={at={(0.55,1.0)},anchor=south,cells={align=left},draw=none,font=\scriptsize,text width=2.5cm},
			legend columns=-1,
			%
			enlarge x limits=0.1,
			%
		]

		\addplot [only marks, bblue, mark=*] table[x=accuracy, y=flops] {./data/benefit_curve_data_KWS_mv.dat};
		\addplot [only marks, ppurple, mark=o] table[x=accuracy, y=flops] {./data/benefit_curve_data_KWS_no_mv.dat};
		\addplot [only marks, black, solid,thick, mark=x] table[x=ref-accuracy, y=ref-flops] {\data};
		 \draw [loosely dashed, red] (92.2,0) -- (92.2,5.54) -- (0,5.54);
		
		
		\addlegendentry{with early-view}
		\addlegendentry{without early-view}

		\end{axis}
	\end{tikzpicture}
        \label{fig:benefit_KWS}
        \caption{DS-CNN}
    \end{subfigure}
    \hspace{0cm}
    \begin{subfigure}[]{0.3\paperwidth}
        \centering
        \begin{tikzpicture}
	\pgfplotstableread{
	ref-accuracy 	ref-flops
	85.81            15.69

	}\data
		\begin{axis}[
			width=\paperwidth/3,
			height=\graphHeight*2- 1.8cm,
			%
			major tick length=3pt,
			%
			major grid style={dashed,color=gray!50},
			minor grid style={color=gray!50},
			ymajorgrids=true,
			xmajorgrids=true,
			%
			xticklabel style={text height=5pt,font=\scriptsize},
			%
			xlabel = Accuracy(\%),
			xlabel style={font=\scriptsize},
		    xlabel near ticks,
			%
			yticklabel style={font=\scriptsize},
			%
			ylabel style={align=center,font=\scriptsize},
			ylabel near ticks,
			legend style={at={(0.55,1.0)},anchor=south,cells={align=left},draw=none,font=\scriptsize,text width=2.5cm},
			legend columns=-1,
			%
			enlarge x limits=0.1,
			%
		]

		\addplot [only marks, bblue, mark=*] table[x=accuracy, y=flops] {./data/benefit_curve_data_VWW_mv.dat};
		\addplot [only marks, ppurple, mark=o] table[x=accuracy, y=flops] {./data/benefit_curve_data_VWW_no_mv.dat};
		\addplot [only marks, black, solid,thick, mark=x] table[x=ref-accuracy, y=ref-flops] {\data};
		 \draw [loosely dashed, thick, red] (85.81,0) -- (85.81,15.69) -- (0,15.69);
		
		
		\addlegendentry{with early-view}
		\addlegendentry{without early-view}

		\end{axis}
	\end{tikzpicture}
        \label{fig:benefit_VWW}
        \caption{Mobilenet}
    \end{subfigure}
    }
    \caption{Benefit curve for all evaluated models}
    \label{fig:benefit_curve}
\end{figure*}
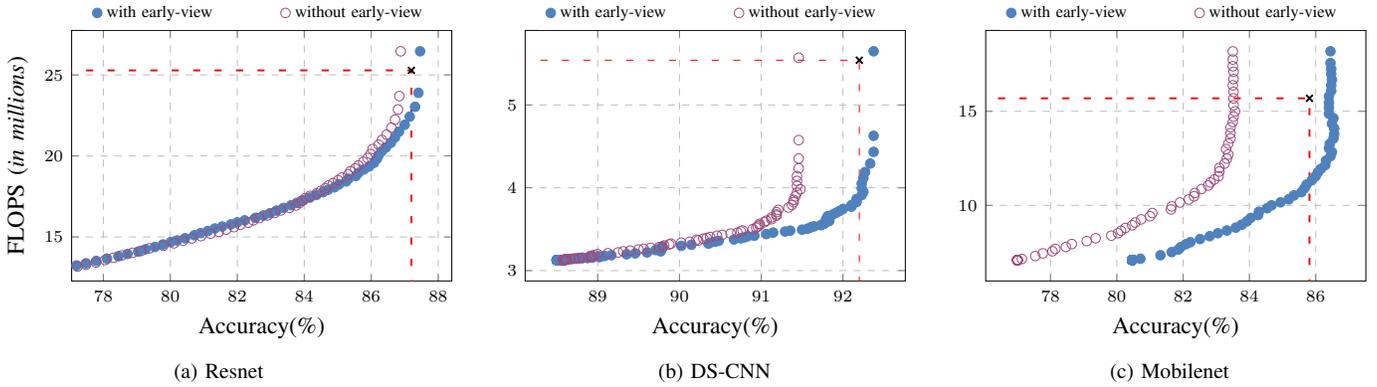

\begin{table*}[tbh]
\centering
\small
\scalebox{0.9}{

{

\begin{tabular}{|c | c | c | c | c | c | c | c | c | c | }
\hline
 \multirow{3}{*}{{\textbf{Model}}} &  \multicolumn{3}{c|}{{\textbf{Baseline}}}  & \multicolumn{6}{c|}{{\textbf{\tool}}} \\   
 
\cline{2-10}
 & \multirow{2}{*}{\textbf{Accuracy}} & {\textbf{FLOPS}} & \multirow{2}{*}{\textbf{Params}} &\multicolumn{3}{c|}{\textbf{FLOPS (millions) at}} & \multirow{2}{*}{\textbf{Params}} & \multicolumn{2}{c|}{\textbf{Acc\textsubscript{standalone}}} \\   

 & & (milions)& & 1\% tradeoff & 2\% tradeoff & 3\% tradeoff & & \textbf{E-e}& \textbf{E-f} \\
  \hline
\rule{0pt}{10pt}Resnet (IC) & 87.2\% & 25.28 & 78.7K & 20.04 & 18.43 & 17.59 & 89.7K & 73.3\% & \textbf{87.4}\%\\
\hline
\rule{0pt}{10pt}DS-CNN (KWS) & 92.2\% & 5.54 & 24.9K & 3.45 & 3.31 & 3.18 & 28.2K & 88.5\% & \textbf{92.37}\%\\
\hline
\rule{0pt}{10pt}Mobilenetv1 (VWW) & 85.81\% & 15.69 & 221.7K & 9.99 & 8.97 & 8.36 & 226.2K & 80.46\% & \textbf{86.44}\%\\
 \hline
\end{tabular}
}
}
\caption{\centering Comparison of \tool with baseline network}
\label{table:accuracy_vs_flops}
\end{table*}

\section{Evaluation Methodology}
\label{sec:methodology}
We evaluate \tool on three tasks from the MLPerf tiny benchmark suite~\cite{tiny-benchmark} published by \emph{MLCommons}: image classification, keyword spotting and visual wake words. These models are tailor-made for tinyML. Further, we evaluate with the exact training scripts (and hyperparamters) used by MLPerf tiny on their official github page~\cite{tiny-git}. The frameworks used are Tensorflow v2.3 with Python v3.7 and models are trained on a GeForce RTX 2080 GPU. 

\noindent\textbf{Image classification}~The image classification task uses a customized Resnet~\cite{tiny-benchmark} that classifies images into ten categories (class) of the CIFAR-10 dataset~\cite{cifar-10}. The model is trained for 500 epochs with a mini-batch size of 32. The Resnet model has 3 residual stacks. We place the early exit immediately after the 1\textsuperscript{st} residual stack.  

\noindent\textbf{Keyword spotting}~Keyword spotting recognizes a short phrase uttered by the user. The model used for the task is a depthwise-separable CNN (DS-CNN)~\cite{DS-CNN} with 4 layers of depthwise separable convolutions. It distinguishes between 12 different classes on the Speech command dataset~\cite{speech_commands}: 10 words $+$ \emph{silence} + \emph{unknown} class. The model is trained for 36 epochs with a mini-batch size of 100. The early-exit is placed immediately after the 2\textsuperscript{nd} depthwise separable layer (i.e. at $\frac{1}{2}\textsuperscript{th}$ of the network).

\noindent\textbf{Visual wake words}~The visual wake words is a binary classification problem that detects the presence of a person if the person occupies more than 2.5\% of the input image. The model used is MobilenetV1~\cite{mobilenets}, which outputs two classes: person and no person. The model is evaluated on the visual wake words dataset~\cite{vww_dataset} derived from the MSCOCO 2014 dataset~\cite{coco}. For evaluation, we use the COCO minival dataset~\cite{minival}~\cite{coco-val2017}. The Mobilenet model is trained for 50 epochs with a mini-batch size of 32. The mobilenet model has 14 layers; the early-exit is placed 
after the 4\textsuperscript{th} layer.

The value of $w\textsubscript{E-e}$ used in our experiments that gives the maximum standalone accuracy at the final exit are: Resnet:$0.5$, DS-CNN:$0.3$ and Mobilenet:$0.4$ (see Section~\ref{subsec:arch-ee}).

\section{Results}
\label{sec:results}
\tool achieves an average of \textbf{31.58\%} reduction in FLOPS (inference time) in exchange for one percent accuracy across all evaluated models (Section~\ref{subsec:results_benefit_curve}) with $\sim9\%$ increase in model size on average. The FLOPS are obtained using~\cite{keras-flops}. Further, we show that the proposed techniques enhance the classification accuracy of the baseline network in all models that we evaluate. Section~\ref{subsec:ee-effectiveness} compares the performance of \tool's early-exit block with the baseline. Next, in Section~\ref{subsec:results-ee-view} we compare the effectiveness of early-view assisted classification and discuss its role in increasing the accuracy of the final classifier. Finally, Section~\ref{subsec:comp_prior} shows that \tool consistently outperforms prior methods. The proposed early-exit strategy is orthogonal to other energy reduction techniques such as pruning and quantization, and can be used in conjunction with them. For example, applying post-training \textsc{int8} quantization on the proposed early-exit equipped Mobilenet model results in 70.4\% reduction in model size. 

\subsection{Accuracy vs Inference cost (FLOPS)}
\label{subsec:results_benefit_curve}
Figure~\ref{fig:benefit_curve} plots the FLOPS consumed as a function of the total accuracy (Equation~\ref{eq:acc_total}), averaged across all test samples. We refer to this plot as the \emph{benefit curve}. All plots are obtained by varying $\rho$ from $0$ to $1$ in step sizes of $0.01$ (see Eq~\ref{eq:model_ee}). As $\rho$ decreases, \Vsubscre{EE}{rate} increases leading to a steady FLOPS reduction. However, the accuracy starts to drop quickly with an increase in \Vsubscre{EE}{rate}. This effect is illustrated in the benefit curves.  Figure~\ref{fig:benefit_curve}  plots the benefit curve for the evaluated networks with and without early-view assisted classification. The accuracy and the FLOPS of the baseline model is marked by '$\times$' in each plot. For higher values of $\rho$, the accuracy of the network is maintained (better even in all cases) close to the baseline accuracy. 
Table~\ref{table:accuracy_vs_flops} reports the improvement in the average FLOPS as $\rho$ reduces. We report the accuracy and the FLOPS at three trade-off points. The trade-off points corresponds to the reduction in FLOPS obtained for every percent of accuracy sacrificed. For example, for the Resnet model with a baseline accuracy of 87.2\%, we report the FLOPS at the points closest to 86.2\%, 85.2\% and 84.2\%.

\vspace{1mm}\noindent
\textbf{Image Classification (IC)}~For the image classification (IC) task, the Resnet model achieves a 20.7\% reduction in FLOPs on average at the first trade-off point, and an additional FLOPS reduction of 6.39\% and 9.71\% respectively at the next two trade-off points.
As seen in Figure~\ref{fig:benefit_curve} and Table~\ref{table:accuracy_vs_flops}, the maximum benefit is gained in exchange for the first percent of accuracy. The improvement in inference time far outweighs the accuracy reduction in this region. 
In addition to this, \tool attains a higher accuracy than that of the baseline. It delivers around 8.9\% reduction in the average FLOPS even before the accuracy starts to fall below that of the baseline. 
The ideal benefit curve has a steep slope. The longer the curve maintains a steep slope, the higher is the reduction in FLOPS without compromising accuracy. The slope of the benefit curve for the Resnet model is affected by two factors. First, there is some disparity in the learning capacities of the early-exit and the final exit. The standalone accuracy of the early-exit i.e. when all inputs exit at the early-exit is 73.39\%. On the other hand, the standalone accuracy of the final exit (i.e. no early-exit) is 87.46\%.
As a result, as $\rho$ decreases more input samples exit at the early-exit and the overall accuracy is impacted by higher misclassifications at the early-exit. 
Second, the image classification task on the CIFAR-10 dataset is a 10-class classification problem. We observe that as the number of classes increases the \emph{confusion} (misclassification rate) of the network rises. We investigate this phenomenon with the Resnet model. In our experiments, we add an extra class to the Resnet model with no changes to either the training data or model hyperparameters. We observe that the standalone accuracies of the early and final exit drop by 0.54\% and 0.75\% respectively.     
This observation is consistent with the other models we evaluate. 
The proposed techniques mitigate the confusion at the early-exit to an extent by instilling higher confidence in the predictions of the early-exit classifier. Section~\ref{subsec:ee-effectiveness} presents more details.

\vspace{1mm}\noindent
\textbf{Keyword spotting (KWS)}~The benefit curve of DS-CNN model~\cite{DS-CNN} for keyword spotting sees the highest reduction in FLOPS (37.7\% reduction) in the one-percent trade-off region. At the next two trade-off points, the FLOPS consumed by the network falls down to 40.2\% and 42.5\%. 
Furthermore, the standalone accuracies of the early-exit and final exit are 88.5\% and 92.37\% respectively. Like Resnet, the standalone accuracy of the final classifier exceeds that of the baseline network, and the total accuracy remains very close to the baseline accuracy for the most part. The model delivers the same network accuracy as that of the baseline network while consuming just 69.6\% of the FLOPS compared to the baseline. 

The lower difference in the standalone accuracies of the two exits in DS-CNN compared to Resnet is because the early and the final exits are separated by just two depthwise separable convolutions. Thus, the proximity in the learning capacities of the two exits ensures that as the $EE\textsubscript{rate}$ increases the early-exit classifier reliably classifies the input samples in an equivalent manner to that of the final classifier in the 1\% tradeoff region thereby, leading to a steep benefit curve. Beyond this region, as the $EE\textsubscript{rate}$ increases, the disparity in the model capacities of the early and final exit becomes more prominent leading to steady accuracy degradation.

\vspace{1mm}\noindent
\textbf{Visual wake word detection (VWW)}~The Mobilenet model for visual wake word detection possesses the best benefit curve compared to the other models we evaluate. For the most part, the curve lies beyond the baseline network's accuracy while consistently delivering reduction in FLOPS. Also, the slope remains steep for the majority of the curve. In fact, it delivers a peak reduction of 29.2\% in FLOPS without falling below the baseline's accuracy. Our investigations revealed that since the early and the final exit are separated by 10 depthwise separable convolution layers, the problem of network overthinking is more prominent in Mobilenet. Therefore, the early-view (EV) assisted classification technique delivers higher returns. This demonstrates the effectiveness of EV-assitance in mitigating overthinking, potentially for even large CNNs. 
Further, the reduction in the average FLOPS consumed is 36.3\% at the first trade-off point and 42.8\% and 46.7\% at the next two trade-off points respectively. 
The standalone accuracies of the early and final exit for Mobilenet are 80.46\% and 86.44\% respectively. 
In addition to this, there is a more gradual reduction of FLOPS as $\rho$ decreases compared to the abrupt fall we observe in DS-CNN.
Our analysis showed that this behavior is because visual wake word detection is a binary classification task. The classification criteria for a binary classification is when either of the class scores is $\geq0.5$. Therefore, there is less scope for confusion. As a result, although the confidence of the early-exit classifier lowers progressively as $\rho$ reduces, its effect on the accuracy only becomes evident much later. Hence, despite a 10-layer separation between the two exits, the standalone accuracies are closer than expected. 
Thus, the slope of the benefit curve remains steep for large parts.

\begin{figure}[t]
\begin{center}
\vspace{-0.5cm}
\scalebox{0.8}{
\includegraphics[height=\graphHeight*2-0.8cm,width=\columnwidth,angle=0]{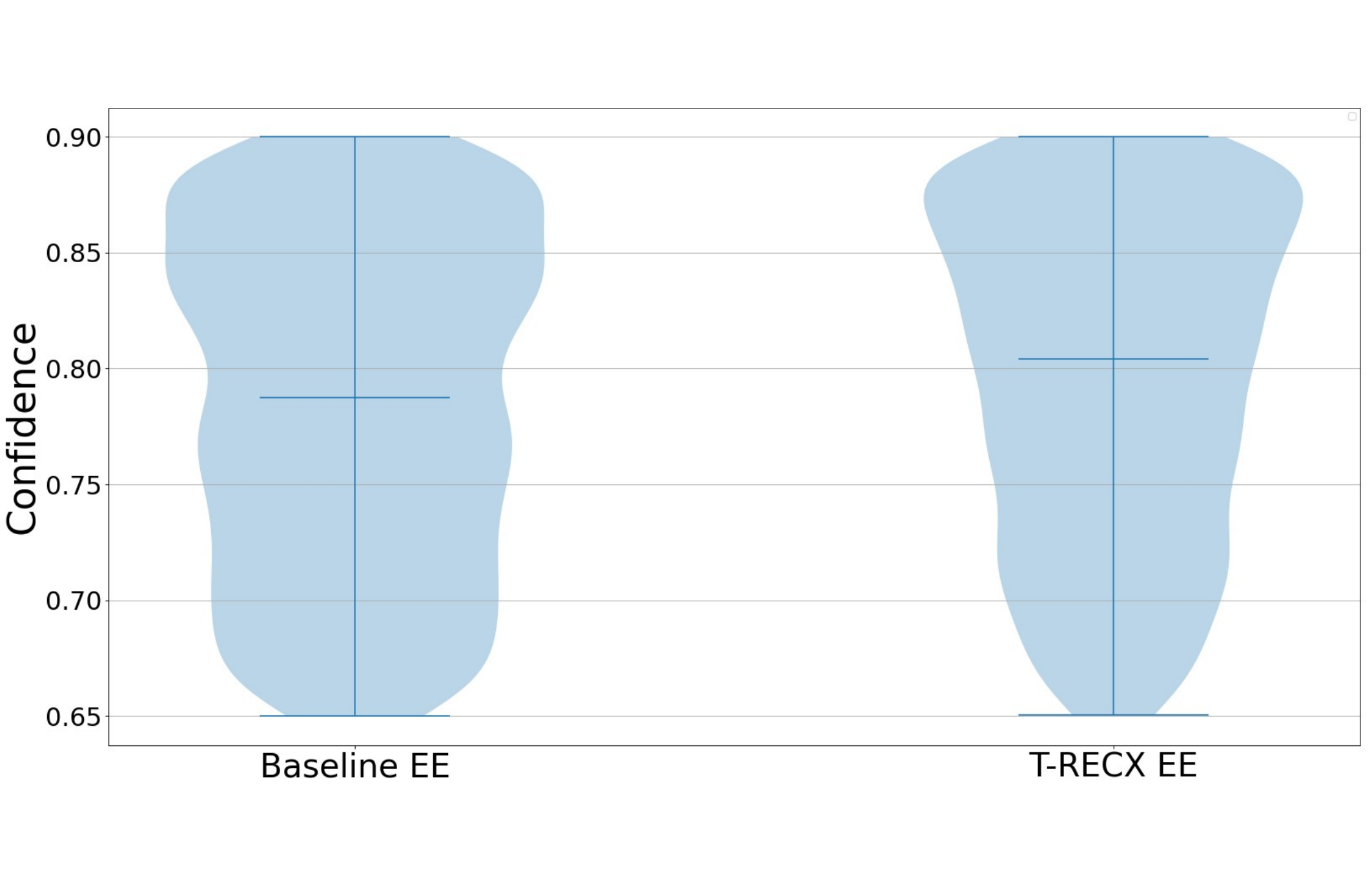}
}
\vspace{-0.0cm}
\centering\caption{Violin Plot of medium confidence predictions ($0.65\leq max(softmax scores)<0.9$)}\vspace{0cm}
\label{fig:violin_plot}
\end{center}
\end{figure}

\subsection{Effectiveness of \tool's early-exit block}
\label{subsec:ee-effectiveness}
As discussed in Section~\ref{sec:related}, in most of the prior works, the early-exit block has either 1)~a simple feature reduction stage (pooling)~\cite{sdn, msdnet, class-spec} or 2)~more learning layers such as convolutions and dense layers followed by feature reduction~\cite{branchynet, conv-at-ee}. In both cases, a final dense layer is included for classification. Since most prior works attach multiple early-exits, they opt for a minimal design with only pooling like the former case. We label this version of the early-exit block as \emph{baseline EE}. However, we find that this design is untenable for tiny-CNNs with single early-exit. We demonstrate the effectiveness of \tool's early-exit block with respect to baseline EE.
Figure~\ref{fig:violin_plot} shows the confidence distribution of all predictions obtained at the early-exit for the Resnet model for two configurations: 1)~with \tool's early-exit block and 2)~with baseline EE block. The y-axis plots the maximum score for each prediction. We consider predictions that lie in the medium confidence region ($0.65\leq max(softmax scores) <0.9$). We present the medium confidence region because we find that there is no discernible change in the network's accuracy for scores $>0.9$. Similarly, we do not present the low confidence region ($max(softmaxscores)<0.65$) because the accuracy degradation in this region makes it impractical. As shown in Figure~\ref{fig:violin_plot}, the violin plot of \tool's early-exit block is noticeably leaner below the median line compared to that of the baseline. Also, the median line for \tool's early-exit block is higher than that of the baseline EE. Evidently, the confidence of the predictions with \tool's early-exit block is notably higher than that of baseline EE. This results in a higher \Vsubscre{EE}{rate} with lower misclassifications at the early-exit. Nevertheless, to put it in context, we present the results obtained on Resnet-8, which produces the least performance improvement out of all models we evaluate.

\subsection{Effectiveness of Early-view assistance}
\label{subsec:results-ee-view}
Figure~\ref{fig:benefit_curve} also plots the benefit curve for all the models we evaluate without early-view (EV) assisted classification. As seen in the figure, the EV-assisted classification outperforms the non-EV versions for all evaluated models. This underlines the effectiveness of the proposed technique in addressing the problem of network overthinking. 
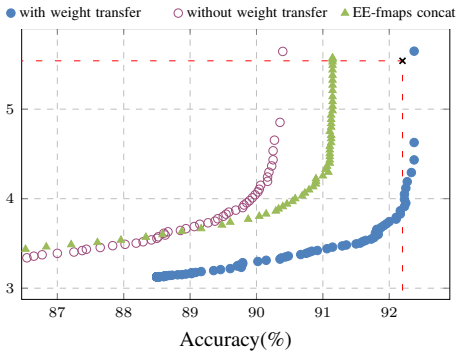
\begin{figure}[tb]
    \centering
    \scalebox{0.8}{\begin{tikzpicture}
	\pgfplotstableread{
	ref-accuracy 	ref-flops
	92.2            5.54

	}\data
		\begin{axis}[
			width=(\paperwidth/3)+1.5cm,
			height=\graphHeight*2 -1cm,
			%
			major tick length=3pt,
			%
			major grid style={dashed,color=gray!50},
			minor grid style={color=gray!50},
			ymajorgrids=true,
			xmajorgrids=true,
			%
			xticklabel style={text height=5pt,font=\scriptsize},
			%
			xlabel = Accuracy(\%),
			xlabel style={font=\scriptsize},
		    xlabel near ticks,
                xmin=87,
			%
			yticklabel style={font=\scriptsize},
			%
			ylabel style={align=center,font=\scriptsize},
			ylabel near ticks,
			legend style={at={(0.55,1.0)},anchor=south,cells={align=left},draw=none,font=\scriptsize,text width=2.5cm},
			legend columns=-1,
			%
			enlarge x limits=0.1,
			%
		]

		\addplot [only marks, bblue, mark=*] table[x=accuracy, y=flops] {./data/benefit_curve_data_KWS_mv.dat};
		\addplot [only marks, ppurple, mark=o] table[x=accuracy, y=flops] {./data/benefit_curve_data_KWS_mv_no_transfer.dat};
            \addplot [only marks, ggreen, mark=triangle*] table[x=accuracy, y=flops] {./data/benefit_curve_data_KWS_msd.dat};
		\addplot [only marks, black, solid,thick, mark=x] table[x=ref-accuracy, y=ref-flops] {\data};
		 \draw [loosely dashed, red] (92.2,0) -- (92.2,5.54) -- (0,5.54);
		
		
		\addlegendentry{with weight transfer}
		\addlegendentry{without weight transfer}
            \addlegendentry{EE-fmaps concat}

		\end{axis}
	\end{tikzpicture}}
    \caption{Benefit curve for DS-CNN 1)~w/ weight transfer 2)~w/o weight transfer from early to final exit, and 3)~concatenation of early-exit and final exit feature maps at final exit}\vspace{-0.3cm}
    \label{fig:benefit_curve_ev}
\end{figure}
For all models, we observe that the EV-assistance at the final exit results in significant improvement in the standalone accuracy of the final exit compared to non-EV-assisted versions i.e. $+0.6\%$ for Resnet, $+0.9\%$ for DS-CNN and $+2.9\%$ for Mobilenet. This in turn ensures higher FLOPS reduction before the curve hits the trade-off points. 
Furthermore, to demonstrate the significance of knowledge transfer from the early-exit to the final exit, we evaluate the DS-CNN model with and without weight transfer using the architecture shown in Figure~\ref{fig:early-view}. In addition, like Huang et al.~\cite{msdnet} and Bonato et al.~\cite{class-spec}, we also evaluate a configuration that concatenates the feature maps of the early-exit with that of the final exit (denoted by \emph{EE-fmaps concat} in Figure~\ref{fig:benefit_curve_ev}). The benefit curves for all three configurations are plotted in Figure~\ref{fig:benefit_curve_ev}. As seen in Figure~\ref{fig:benefit_curve_ev}, the model \emph{with weight transfer} comfortably outperforms the version without weight transfer with an average accuracy improvement of $+1.98\%$. This clearly illustrates that the accuracy improvement is due to EV-assistance, and not simply because of additional learning layers. Interestingly, the model with concatenation of early and final exit feature maps also outperforms the version without weight transfer, which indicates the presence of \emph{overthinking} and the effectiveness of leveraging the knowledge of early-exit to alleviate its effect on accuracy. However, as seen from Figure~\ref{fig:benefit_curve_ev}, it is evident that EV-assistance (weight transfer) method is more effectual compared to na\"{\i}ve concatenation of feature maps. 

\subsection{Comparison with prior works}
\label{subsec:comp_prior}
Figure~\ref{fig:benefit_curve_prior} compares \tool with two popular prior works: Shallow Deep Networks (SDN)~\cite{sdn} and Branchynet~\cite{branchynet}. As discussed in Sections~\ref{sec:related} and~\ref{sec:motivation}, both Branchynet and SDN focus on large ML models and employ multiple early-exits. The former places additional convolution layers at its two early-exits before pooling+dense, and the latter opts for higher number of early-exits with a minimalist design of pooling+dense at all exits. We apply the early-exit techniques presented in Branchynet and SDN on two tinyCNNs we evaluate: Resnet-8 and DSCNN. Due to space constraints we omit the results for MobilenetV1 because like DS-CNN, it is also a depthwise-separable model. We denote the branchynet versions of Resnet and DSCNN as \bresnet and \bdscnn, and their SDN versions as \sresnet and \sdscnn respectively. 
\bresnet and \bdscnn employ two early-exits at $\frac{1}{3}$\textsuperscript{rd} and $\frac{2}{3}$\textsuperscript{rd} of its networks. The early-exit blocks of \bresnet and \bdscnn have an additional convolution layer before the pooling+dense layers. On the other hand, \sresnet uses two early-exits (after each residual stack) and \sdscnn uses three early-exits (after each depthwise-separable layer) respectively. Both \sresnet and \sdscnn employ a simple pooling+dense configuration at each of their early exits. 
\begin{figure}[t!]
    \centering
    \scalebox{0.8}{
    \hspace{-6mm}
    \begin{subfigure}[t]{0.3\paperwidth}
        \centering
        \begin{tikzpicture}
	\pgfplotstableread{
	ref-accuracy 	ref-flops
	87.2            25.28

	}\data
		\begin{semilogyaxis}[
			width=\paperwidth/3.4,
			height=\graphHeight*2 -1cm,
			%
			major tick length=3pt,
			%
			major grid style={dashed,color=gray!50},
			minor grid style={color=gray!50},
			ymajorgrids=true,
			xmajorgrids=true,
			%
			xticklabel style={text height=5pt,font=\scriptsize},
			%
			xlabel = Accuracy(\%),
			xlabel style={font=\scriptsize},
		    xlabel near ticks,
                xmin=75,
			%
			yticklabel style={font=\scriptsize},
			%
			ylabel style={align=center,font=\scriptsize},
			ylabel=FLOPS (\textit{in millions}),
			ylabel near ticks,
			legend style={at={(0.55,1.0)},anchor=south,cells={align=left},draw=none,font=\scriptsize,text width=1.5cm},
			legend columns=-1,
			%
			enlarge x limits=0.1,
			%
		]

		\addplot [only marks, bblue, mark=*] table[x=accuracy, y=flops] {./data/benefit_curve_data_IC_mv.dat};
		\addplot [only marks, ggreen, mark=triangle*] table[x=accuracy, y=flops] {./data/benefit_curve_data_IC_branchynet.dat};
            \addplot [only marks, yyellow, mark=o] table[x=accuracy, y=flops] {./data/benefit_curve_data_IC_sdn.dat};
		\addplot [only marks, black, solid,thick, mark=x] table[x=ref-accuracy, y=ref-flops] {\data};
		 \draw [loosely dashed, thick, red] (87.2,0) -- (87.2,25.28) -- (0,25.28);
		
		
		\addlegendentry{\tool}
		\addlegendentry{Branchynet}
            \addlegendentry{SDN}

        
		\end{semilogyaxis}
	\end{tikzpicture}
        \label{fig:benefit_prior_IC}
        \caption{Resnet}
    \end{subfigure}
    \hspace{-0.7cm}
    \begin{subfigure}[t]{0.3\paperwidth}
        \centering
        \begin{tikzpicture}
	\pgfplotstableread{
	ref-accuracy 	ref-flops
	92.2            5.54

	}\data
		\begin{semilogyaxis}[
			width=\paperwidth/3.4,
			height=\graphHeight*2-1cm,
			%
			major tick length=3pt,
			%
			major grid style={dashed,color=gray!50},
			minor grid style={color=gray!50},
			ymajorgrids=true,
			xmajorgrids=true,
                xmin=88,
			%
			xticklabel style={text height=5pt,font=\scriptsize},
			%
			xlabel = Accuracy(\%),
			xlabel style={font=\scriptsize},
		    xlabel near ticks,
			%
			yticklabel style={font=\scriptsize},
			%
			ylabel style={align=center,font=\scriptsize},
			ylabel near ticks,
			legend style={at={(0.55,1.0)},anchor=south,cells={align=left},draw=none,font=\scriptsize,text width=1.5cm},
			legend columns=-1,
			%
			enlarge x limits=0.1,
			%
		]

		\addplot [only marks, bblue, mark=*] table[x=accuracy, y=flops] {./data/benefit_curve_data_KWS_mv.dat};
		\addplot [only marks, ggreen, mark=triangle*] table[x=accuracy, y=flops] {./data/benefit_curve_data_KWS_branchynet_div2.dat};
            \addplot [only marks, yyellow, mark=o] table[x=accuracy, y=flops] {./data/benefit_curve_data_KWS_sdn.dat};
		\addplot [only marks, black, solid,thick, mark=x] table[x=ref-accuracy, y=ref-flops] {\data};
		 \draw [loosely dashed, red] (92.2,0) -- (92.2,5.54) -- (0,5.54);
		
		
		\addlegendentry{\tool}
            \addlegendentry{Branchynet}
            \addlegendentry{SDN}

		\end{semilogyaxis}
	\end{tikzpicture}
        \label{fig:benefit_prior_KWS}
        \caption{DS-CNN}
    \end{subfigure}
    }
    \caption{Benefit curves of \tool, Branchynet~\cite{branchynet} and Shallow Deep Networks (SDN)~\cite{sdn} (y-axis in log-scale)}
    \label{fig:benefit_curve_prior}
\end{figure}
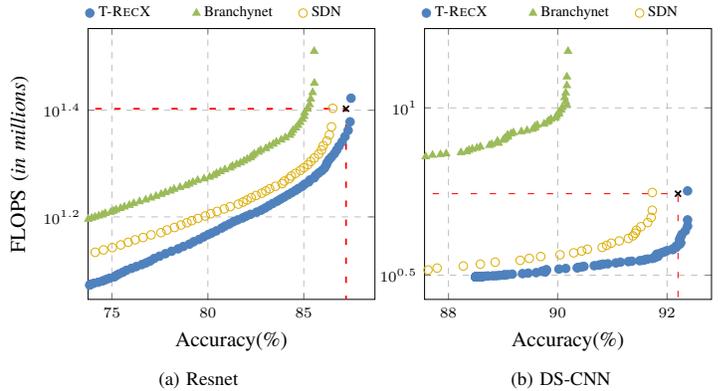
\begin{table}[t!]
\centering
\scalebox{0.73}{
{
\begin{tabular}{|c | c | c | c | c | c | c | c | c | c | }
\hline
 \multirow{3}{*}{\textbf{Model}} & \multicolumn{3}{c|}{\textbf{\tool}} & \multicolumn{3}{c|}{\textbf{Branchynet}} & \multicolumn{3}{c|}{\textbf{SDN}}\\
 \cline{2-10}
  & \multirow{1}{*}{\textbf{Par-}} & \multicolumn{2}{c|}{\textbf{Accuracy(\%)}} & \multirow{1}{*}{\textbf{Par-}} & \multicolumn{2}{c|}{\textbf{Accuracy(\%)}} & \multirow{1}{*}{\textbf{Par-}} & \multicolumn{2}{c|}{\textbf{Accuracy(\%)}}\\  
   & \textbf{ams} & \textbf{E-e}& \textbf{E-f} & \textbf{ams}& \textbf{E-e} & \textbf{E-f} & \textbf{ams}& \textbf{E-e} & \textbf{E-f}\\
   \hline
   \textbf{Resnet} & 89.7K & 73.4 & \textbf{87.4} & 98.4K & 72.4, 79.7 & 85.7 & 94K & 63.3, 75 & 86.3\\ 
   \textbf{DSCNN} & 28.2K & 88.8 & \textbf{92.3} & 62.6K & 85, 87.6 & 90.9 & 38.7K & 63.7, 81.7, 89 & 91.8 \\
   \hline
\end{tabular}
}
}
\caption{\centering Comparison of \tool with Branchynet and SDN. (Accuracy numbers reported here are standalone accuracies)\vspace{0mm}}
\label{table:comp_prior}
\end{table}



 


Figure~\ref{fig:benefit_curve_prior} compares all benefit curves. 
\tool consistently outperforms both Branchynet and SDN. The model parameters of \tool are on average \emph{$31.8\%$ smaller} than Branchynet and \emph{$15.8\%$ smaller} than SDN. Further, Branchynet consumes significantly higher FLOPS compared to other methods because of an additional CONV layer at its early-exits. This is especially noticeable in DSCNN because the CONV layers are more compute/memory intensive compared to depthwise separable convolutions. 
Also, we observe that the benefit curves of both Branchynet networks are similar to the non-EV (no weight transfer) versions of Figure~\ref{fig:benefit_curve}. This result shows that simply adding additional layers at early-exits is not sufficient for accuracy preservation in tiny-CNNs, which in turn exhibits the value of EV-assistance. Furthermore, SDN networks outperform the Branchynet versions because they have a higher number of early-exits. However, the benefit curves of \tool is still comfortably superior compared to SDN. Table~\ref{table:comp_prior} reports the model sizes and the standalone accuracies of all early-exits and that of the final exit for all evaluated models. In particular, \tool achieves the highest standalone accuracy at the final exit due to EV-assistance, which ensures steady FLOPS reduction with minimal compromise on accuracy. 
\section{Conclusions}
\label{sec:conclusion}
Existing works on early-exit networks either target large networks making them inapplicable for tiny-CNNs or optimize solely for energy while compromising heavily on accuracy. Further, tiny-CNNs have higher sensitivity to early-exits compared to large CNNs. This paper addresses these issues by 1) presenting a novel early-exit architecture for tiny-CNNs, 2) describing a joint training method that mitigates the disruption of neurons relations. Additionally, we show a technique to mitigate the destructive effect of network overthinking at the final exit by distilling knowledge from the early-exit block. \tool achieves an average of 31.58\% reduction in FLOPS by trading one percent of accuracy across all evaluated models and consistently outperforms prior works. Also, the proposed techniques increase the accuracy of the baseline network in all models we evaluate. The methods presented in this work can easily be extended to large CNNs as well. \tool provides a simple way to trade-off between accuracy and inference time (FLOPS) by tuning a single parameter either pre/post deployment based on compute/energy budget, which is useful for the tinyML space to manage battery life.

\printbibliography
\end{document}